\documentclass[10pt,twocolumn,letterpaper]{article}

\usepackage{iccv}
\usepackage{times}
\usepackage{epsfig}
\usepackage{graphicx}
\usepackage{amsmath}
\usepackage{amssymb}
\usepackage{float}

\usepackage[breaklinks=true,bookmarks=false]{hyperref}

\iccvfinalcopy 

\def\iccvPaperID{\idpaper} 

\ificcvfinal\fi

\newcommand{\vct}[1]{\ensuremath{\boldsymbol{#1}}}

\newcommand{\myparagraph}[1]{\noindent \textbf{#1} --}

\newcommand{\settitle}{Adversarial Attacks Against Uncertainty Quantification}
\newcommand{\idpaper}{40}

\newcommand{\ellinf}{$\ell_{\infty}$\xspace}

\begin{document}

\title{\settitle}

\author{{Emanuele Ledda$^{1,3}$, 
Daniele Angioni$^{1,2}$, 
Giorgio Piras$^{1,2}$, 
Giorgio Fumera$^{2}$, 
Battista Biggio$^{2}$, 
Fabio Roli$^{3}$}\\
{\normalsize $^1$Department of Computer, Control and Management Engineering, Sapienza University of Rome, Italy}\\
{\tt \small emanuele.ledda@uniroma1.it}\\
{\normalsize $^2$Department of Electric and Electronic Engineering, University of Cagliari, Italy}\\
{\tt \small \{daniele.angioni, giorgio.piras, fumera, battista.biggio\}@unica.it}\\
{\normalsize $^3$Department of Informatics, Bioengineering, Robotics, and Systems Engineering, University of Genova, Italy}\\
{\tt \small fabio.roli@unige.it} 
}


\maketitle
\ificcvfinal\thispagestyle{empty}\fi

\begin{abstract}
    Machine-learning models can be fooled by adversarial examples, i.e., carefully-crafted input perturbations that force models to output wrong predictions.
    While uncertainty quantification has been recently proposed to detect adversarial inputs, under the assumption that such attacks exhibit a higher prediction uncertainty than pristine data, it has been shown that adaptive attacks specifically aimed at reducing also the uncertainty estimate can easily bypass this defense mechanism.
    In this work, we focus on a different adversarial scenario in which the attacker is still interested in manipulating the uncertainty estimate, but regardless of the correctness of the prediction; in particular, the goal is to undermine the use of machine-learning models when their outputs are consumed by a downstream module or by a human operator.
    Following such direction, we: \textit{(i)} design a threat model for attacks targeting uncertainty quantification; \textit{(ii)}  devise different attack strategies on conceptually different UQ techniques spanning for both classification and semantic segmentation problems;
    \textit{(iii)} conduct a first complete and extensive analysis to compare the differences between some of the most employed UQ approaches under attack.
    Our extensive experimental analysis shows that our attacks are more effective in manipulating uncertainty quantification measures than attacks aimed to also induce misclassifications.

\end{abstract}

\section{Introduction}
Machine Learning (ML) covers nowadays multiple applications, including safety-critical domains such as medical diagnosis, self-driving cars, and video surveillance. 
Leaning towards ML-based systems tailored to cope with such scenarios, the research community also focused on enhancing the \textit{trustworthiness} of such systems.  
In this regard, Uncertainty Quantification (UQ) methods have been fostered throughout the years, establishing themselves as methods capable of assessing the degree of \emph{uncertainty} of the predictions made by an ML-based system~\cite{HullermeierW21}.
Unfortunately, ML models have been found to be susceptible to carefully-crafted input samples aimed at causing wrong predictions, known as \textit{adversarial examples}~\cite{biggio13-ecml,szegedy_intriguing_2014}. 
Several defensive countermeasures have been developed, aiming to build robust models, including \textit{adversarial training}~\cite{madry18-iclr}  and also uncertainty quantification.
In particular, UQ has been proposed as a \emph{defense} technique for adversarial attack \emph{detection} at test time, based on the rationale that attack samples aimed at causing wrong predictions are characterised by high uncertainty. 
However, analogously to other defense techniques, some works have shown that it is indeed possible to generate \textit{adaptive} attacks capable of causing wrong predictions and at the same time of evading detection, in this case by reducing the corresponding uncertainty measure~\cite{Carlini2017,Grosse2019}. 

In this work, we focus on a different adversarial scenario in which the attacker is still interested in manipulating the uncertainty estimate, but regardless of the correctness of the prediction; in particular, the goal is to undermine the use of UQ techniques for ML models when their outputs are consumed by a downstream module or by a human operator. For instance, in the medical domain, a doctor may avail of uncertainty for distinguishing if an ML prediction (\ie, a tumor segmentation) is reliable enough or requires more attention from the doctor. 
Having an estimate about the reliability of the system's predictions would allow a healthcare operator to accurately weigh its time, giving an additional effort when interpreting more uncertain cases. 
Another example is a crowd counting tool that processes in real-time video streams coming from a video surveillance network to support law enforcement agency officers in crowd monitoring. 
Such a system may provide an estimate of the uncertainty of the predicted crowd count (\eg, in terms of a 95\% confidence interval) to make its users aware of the reliability of its predictions. 
This may allow detecting out-of-distribution (OOD) frame sequences (\eg, due to extreme lighting conditions) that are likely to be characterized by high uncertainty, whose corresponding predicted count would be disregarded by the users. 
We argue that, in application scenarios like the ones described above, an attacker may be interested in undermining \emph{only} the UQ component, regardless of the predictions. 
For the medical domain case, \eg, an attacker may target an ML system to increase the uncertainty associated with its predictions, resulting in an unnecessary additional workload for the operator in charge (\eg, evaluating a tumor diagnosis).
On the crowd counting side, lowering the level of uncertainty may lead the LEA operator to think the count is always correct, even in the presence of under-estimated predictions caused, \eg, by inadequate illumination or extreme weather conditions (and, hence, increasing the odds of casualties).

We thus believe that focusing on attacks targeting the sole uncertainty can be highly relevant for safety applications and that a proper understanding of such attacks, to the best of our knowledge, is still missing. 
Consequently, the state of the art lacks a practical implementation and empirical evaluation of UQ techniques under attack. 
In this work, we move the first steps towards this direction by providing the following contributions: 
\begin{itemize}
    \item We design a threat model for attacks targeting UQ and yielding \emph{wrong uncertainty estimates}; 
    \item We develop and implement different attack strategies on conceptually different UQ techniques spanning over both classification and semantic segmentation tasks; 
    \item We conduct a first complete and extensive analysis to compare the differences between some of the most employed UQ approaches under attack. 
\end{itemize}


\section{Background and Related Work}
We summarize here the essential concepts of UQ techniques, adversarial machine learning, and overview existing work on attacks against UQ.

\subsection{Uncertainty Quantification}
In classification problems characterized by a $d$-dimensional feature space $\mathcal X \subseteq \mathbb{R}^d$ and a $L$-dimensional output space $\mathcal Y \subseteq \mathbb{R}^L$ being $L$ the number of classes, an ML-based predictor implements a decision function $f^{\theta} : \mathcal X \mapsto \mathcal Y$ mapping an input vector to an output categorical distribution, where the parameters $\theta$ are obtained by minimizing a given loss function on a training set $\mathcal D$ of $(\vct x, \vct y)$ pairs.
Predictions are subject to two kinds of uncertainty: \textbf{aleatoric} uncertainty (a.k.a. \emph{data} uncertainty), due to the inherent randomness of the class label (i.e., overlapping class-conditional distributions), and \textbf{epistemic} uncertainty (a.k.a. \emph{model} uncertainty), due to a lack of knowledge on the ``correct'' prediction model (such as the DNN's weights), which can be caused, e.g., by a training set that is not entirely representative for a given task. 
UQ techniques aim to associate with each prediction a numerical estimate of its uncertainty~\cite{HullermeierW21}.

\myparagraph{Probabilistic approaches}
Bayesian Neural Networks (BNNs) are a well-known probabilistic model, which naturally allow assessing the uncertainty of their predictions~\cite{MacKay1992}. 
%
They assume a prior $p(\theta)$ over the model's parameters and marginalize over it to compute a predictive distribution on a given training set $\mathcal{D}$ by \textit{Bayesian Model Averaging} (BMA):
\begin{equation}
    \label{eq:marginalization}
     f_{BMA} = p(y|x,\mathcal{D}) = \int_{\theta} p(y|x,\theta) \cdot p(\theta|\mathcal{D}) \textrm{d}\theta
\end{equation}
%
%
Since Eq.~\ref{eq:marginalization} is intractable in practice, an approximating distribution $q(\theta)$ is commonly used, minimizing its divergence from the actual distribution.
In this work we focus on two common approximations: Monte-Carlo Dropout and Deep Ensemble.

\textbf{Monte-Carlo (MC) dropout} approximation~\cite{GalG16} consists of activating dropout at test time, either in an ad hoc way~\cite{GalG16} (namely embedded dropout), using the dropout rate found during training (where dropout is also used for regularization), or in a post hoc way~\cite{Loquercio2020,Ledda2023} (namely dropout injection), i.e., on already trained networks. 
An alternative solution, which has been shown capable of outperforming MC dropout, is based on \textbf{Deep Ensembles}~\cite{Lakshminarayanan17}, which trains multiple DNNs starting from random weights and approximate BMA by combining the corresponding predictions obtained from the different instances of $\theta$.
In both cases, for a given sample $\vct x$ one can compute its corresponding uncertainty $\mathcal{U}(\vct x)$ by computing a statistic (e.g., the variance) over the Monte-Carlo predictions.

In addition, we recall many other state-of-the-art Bayesian methods. 
Among them, we can find Concrete Dropout~\cite{ConcreteDropout17} (an improvement of MC-dropout for finding the dropout rate during training), BayesByBackprop~\cite{Blundell15}, and the whole class of Laplace Approximations~\cite{MacKay1992,Kristiadi0H21} (which are one of the most prominent post hoc UQ techniques).

\myparagraph{Deterministic approaches}
A drawback of Bayesian models is their computational cost due to the multiple forward passes required to obtain a point-wise prediction. 
Several deterministic approaches have been proposed to deal with this issue, such as Deterministic Uncertainty Quantification (DUQ)~\cite{van2020uncertainty}, Spectral-normalized Neural Gaussian Process (SNGP)~\cite{Liu2020} and Deep Deterministic Uncertainty (DDU)~\cite{mukhoti2021}.

For instance, for $L$-class classification problems, DUQ learns $L$ centroids in the feature space $\mathcal X$ and, for any input $\vct x$, it returns a $L$-dimensional vector with the distance between the feature vector (defined as $f^\theta(\vct x)$ with abuse of notation) and the centroids, computed using a Radial Basis Function (RBF) kernel. 
The predicted class is the one associated to the closest centroid, and the corresponding distance is interpreted as the uncertainty measure $\mathcal{U}(\vct x)$.

\subsection{Adversarial Machine Learning}
ML models have been found to be susceptible to adversarial attacks~\cite{szegedy_intriguing_2014}, i.e., input samples carefully crafted to be misclassified.
Several attacks and defenses have been proposed so far. 
Two seminal yet still widely used attack strategies are the Fast-Gradient Sign Method attack (FGSM)~\cite{goodfellow15-iclr} and the Projected Gradient Descent attack (PGD)~\cite{madry18-iclr}. 
Under a ``standard'' untargeted \ellinf threat model with a perturbation budget $\epsilon$, FGSM crafts an adversarial example $\vct{x^*}$ by adding to a given sample $\vct x$ 
an \ellinf norm perturbation of magnitude $\epsilon$,
pointing to the steepest ascent direction of the loss $L$ from the point $\vct x$:
\begin{equation}
    \vct{x^*}=\vct{x}+\epsilon \cdot \operatorname{sgn}\left(\nabla_{\vct x} L\left(f_\theta(\vct x), \vct y\right)\right) \ ,
\end{equation} 
where $\nabla$ denotes the gradient operator.
The PGD attack implements an iterative version of FGSM by projecting after each iteration the obtained perturbation to the feasible domain $\Gamma=\{\vct{x_t} \in \mathcal X:||\vct{x_t - x_0}||_{\infty} < \epsilon\}$:
\begin{equation}
    \vct{x_{t+1}}=Proj_\Gamma(\vct{x_t}+\alpha \cdot \operatorname{sgn}\left(\nabla_{\vct{x_t}} L\left(f_\theta(\vct{x_t}), \vct y\right)\right)) ,
\end{equation} 
On the defense side, the par excellence technique is Adversarial Training~\cite{madry18-iclr}: 
\begin{equation}\label{eq:adv_train}
    \min _\theta \mathbb{E}_{(\vct x, \vct y) \sim \mathcal{D}}\left[\max _{\delta \in B(\vct x, \varepsilon)} \mathcal{L}(\theta, \vct x+\vct \delta, \vct y)\right] \ ,
\end{equation}
where $B(\vct x, \varepsilon)$ denotes the set of allowed adversarial perturbations, bounded by $\epsilon$. Eq.~\ref{eq:adv_train} amounts to solve a min-max optimization problem, where the worst-case loss $\mathcal{L}$ (inner problem) has to be minimized (outer problem). The goal is to train the model to be robust to adversarial examples.

\subsection{Evasion Attacks Involving Uncertainty}
Previous work in the adversarial machine learning field has considered UQ only as a defense strategy, as a means for detecting adversarial samples crafted for \emph{evading} a classifier, i.e., to cause wrong predictions.
For instance, the authors of~\cite{Feinman2017} proposed to assess uncertainty as the variance computed using embedded MC dropout (with a dropout rate of $0.5$ after each convolutional layer). 
Using a detection threshold $\tau=0.02$, such that samples whose variance is below it are rejected as adversarial, 96\% of adversarial examples were correctly identified and rejected on the CIFAR-10 dataset, with a false-positive rate of 1\%.
Following the usual arms race approach, subsequent works devised evasion attacks capable of bypassing uncertainty-based defenses.
The attack presented in~\cite{Carlini2017} manipulates a given sample to reduce the corresponding MC sample variance below the detection threshold and consequently induces a misclassification.
On the same CIFAR-10 dataset, it bypassed the above defense with a success rate of 98\%. However, this result was attained at the expense of a notably large perturbation size.

The authors of~\cite{Grosse2019}, proposed the ``High-Confidence Low-Uncertainty'' attack. 
For a given sample $x$, the underlying idea is to craft an adversarial example $x+\delta$ pushing the prediction confidence for the target (wrong) class over 95\% and simultaneously keeping the corresponding uncertainty not higher than the one of the original sample. 

Previous work involving UQ considered only \emph{evasion} attacks aimed at causing \emph{wrong predictions}, where uncertainty measures were used and manipulated only as detection tools. 
In this work, we focus instead on a different attack scenario where the goal is to manipulate uncertainty measures \emph{per se}, i.e., to produce \emph{wrong uncertainty estimates}, thus undermining their original purpose of providing an assessment of the reliability of ML-based systems predictions, to be used by a downstream processing module or by a human operator, regardless of the correctness of the predictions. 
Additionally, we extensively test attacks to diverse UQ techniques to assess how such attacks are supposed to mutate depending on the given uncertainty-related scenario.
\section{Uncertainty Quantification Under Attack}
\label{sect:uq_under_atk}

In this section we formally present our threat model, where the attacker's goal is to produce \emph{wrong uncertainty estimates} regardless of the correctness of the prediction, develop a possible implementation for classification tasks, and show how it can be extended to other tasks using semantic segmentation as a case study.

\subsection{Threat Model}
Evasion attacks aim at getting a given sample misclassified, with respect to its ground-truth label.
However, UQ techniques do not have a ground truth, and thus it is not straightforward to define what a ``wrong'' uncertainty estimate is.
Ideally, higher uncertainty values should be associated with higher misclassification probability.
The stronger such statistical correlation is, the more ``correct" the uncertainty measure will get. 
Accordingly, the considered attack against UQ should result in \textit{breaking this statistical correlation up}.

\myparagraph{Taxonomy of attacks to uncertainty quantification}
According to the above threat model, two possible kinds of attacks can be identified: 
\begin{itemize}
    \item Overconfidence Attack (O-attack): Its goal is to \emph{reduce} the uncertainty measure of a given predictor, thus tricking an ML-based system into being overconfident. This will impact in particular wrong predictions and out-of-distribution (OOD) samples, resulting in undermining the UQ module \textit{integrity}. 
    \item Underconfidence Attack (U-attack): The goal of this attack, conversely, is to \emph{increase} the uncertainty measure, which would result in considering all the predictions as unreliable, which in turn would lead the downstream modules or human operators to disregard the outputs of an ML-based system.
    We, therefore, classify the U-attack as a threat undermining the \textit{availability} of an ML-based system.
\end{itemize}

Theoretically, one can formulate the problem as the search for the perturbation $\delta$, bounded by $\epsilon$, minimizing (O-attack) or maximizing (U-Attack) the uncertainty estimate $\mathcal{U}(\vct x + \vct \delta)$:
\begin{equation}\label{eq:uq_attack}
    \underset{\vct \delta}{\operatorname{argmin}} \; \gamma \cdot \mathcal{U}(\vct x + \vct \delta), \quad s.t. \lVert \vct{\delta} \rVert_p < \epsilon \ ,
\end{equation}
where $\gamma \in \{-1,1\}$ controls the attack objective: $\gamma=-1$ corresponds to the U-attack, whereas $\gamma=1$ corresponds to the O-attack.
While the threat model encompasses both attacks, in the rest of our work we focus on the O-Attack, being the latter (just like ``standard" evasion attacks~\cite{biggio18})
a violation of the integrity of the ML-based system.  
Therefore, in the following section, we propose a possible implementation of the O-Attack. 

\subsection{Attacking Probabilistic Models}
\label{sect:atk_prob_models}

The attack strategy of Eq.~\ref{eq:uq_attack} can be implemented both for probabilistic and deterministic UQ models. 

\myparagraph{Minimum Variance Attack}
In probabilistic models, the prediction and uncertainty value for a given sample are obtained by combining a set of predictions. 
Such models commonly leverage uncertainty measures such as predictive variance (epistemic), entropy (aleatoric) or, less frequently, mutual information~\cite{SmithGal2018} (either epistemic or predictive). 
The intrinsic probabilistic nature of such methods requires attacks to rely on expectations over a set of MC samples. 
In this context, a first possible solution consists of modifying a given input sample $x$ in such a way that the predictor's probabilistic outcomes are as \emph{concordant} as possible. This can be formulated as a direct minimization of the predictive variance; 
accordingly, we refer to this attack as Minimum Variance Attack (MVA):
%
%
\begin{equation} \label{eq:mva}
\begin{split}
\underset{\vct \delta}{\operatorname{argmin}}  \; \mathbb{E}_S  [(\vct{x+\delta})^2] & - \mathbb{E}_S [(\vct{x+\delta})]^2, \quad s.t. \lVert \vct{\delta} \rVert_p < \epsilon, \ \\
\mathbb{E}_S [(\vct{x+\delta})^2] := &\frac{1}{S}\sum_{s=1}^S f^{\theta_s}(\vct{x+\delta})^\intercal \cdot f^{\theta_s}(\vct{x+\delta}),\\
\mathbb{E}_S [(\vct{x+\delta})]^2 := &\mathbb{E}_S(\vct{x+\delta})^\intercal \cdot \mathbb{E}_S(\vct{x+\delta}) ,
\end{split}
\end{equation}
where $\vct \delta$ denotes the perturbation, $S$ the Monte-Carlo sample size, and $f^{\theta_s}$ the predictor corresponding to the parameters $\theta_s$ obtained from the $s$-th Monte Carlo sample (see Sect.~\ref{eq:marginalization}).
Finally, $\mathbb{E}_S(\vct x+ \vct \delta) \approx f_{BMA}(\vct  x+ \vct \delta)$ is the Monte-Carlo approximation of the BMA using the set of size $S$.

\myparagraph{Auto-Target Attack}
Albeit the attack described above aims at minimizing the uncertainty measure directly, there are other ways to optimize Eq.~\eqref{eq:uq_attack}.
A simple yet effective alternative idea has indeed been proposed in~\cite{Carlini2017} to evade the detection of adversarial examples (modeled as an uncertainty threshold). 
To this aim, the authors proposed to get the probabilistic model's average prediction closer to the most likely incorrect class; since it is equivalent to choosing an automatic target, we refer to this attack as Auto-Target Attack (ATA). 
A possible formulation can be obtained by minimizing the Cross-Entropy (CE) loss~\cite{Carlini2017}:
\begin{equation}\label{eq:ata_carlini}
    \underset{\vct \delta}{\operatorname{argmin}} \; - \log(\mathbb{E}_S(\vct{x+\delta})_c) \ , \quad s.t. \lVert \vct{\delta} \rVert_p < \epsilon \ ,
\end{equation}
where $c$ denotes the automatically chosen target class and $\mathbb{E}_S(\vct{x+\delta})$ the expectation of the predictions over $S$ Monte-Carlo forward passes.
Bringing the average of a prediction's set closer to a certain target corresponds to getting all the predictions closer to a common target. 
Albeit the above approach was originally formulated as a C\&W attack~\cite{CWAttack}, we point out that it can be extended to several attacks. 

As mentioned in previous work~\cite{Carlini2017}, the above attack required a particularly large perturbation to be effective: such a relatively high perturbation was necessary to evade the model's predictions, besides reducing the uncertainty measure. 
Indeed, using ATA with the primary goal of evading the predictions does not result in a sudden variance minimization but will instead take two stages: 
in the first stage, after a warm-up phase, the variance starts growing as long as the prediction flips from correct to incorrect; in the second stage, the probability of the class being maximized overtakes the others, leading to a further stabilization and, thus, to the variance minimization. 
Therefore, an attacker interested in the efficacy of such an attack should favor correctly classified clean samples over misclassified adversarial examples with higher uncertainty estimates.

\myparagraph{Stabilizing Attack}
We further improve this simple idea by taking the most likely class indiscriminately (instead of the most likely incorrect one) since we are not interested in the correctness of the prediction.
The effect of our formulation, which we name \textit{Stabilizing Attack} (STAB), is to get every MC prediction closer to the mean basin of attraction, thus \emph{stabilizing} the predictions, which results in turn to lower variance and average prediction's entropy.

\subsection{Attacking Deterministic Models}
\label{sect:atk_det_models}
Due to the nature of deterministic models, an attacker can evade their associated uncertainty measure by focusing on a single parameter configuration $\theta$, without the need of MC sampling.
As an example, we show how our STAB attack can be extended to the widely used DUQ technique~\cite{van2020uncertainty} and other deterministic methods.
For a deterministic UQ model, it is sufficient to craft the adversarial sample $\vct{x^*}$ to make it approach a centroid $\vct e_c$ associated to a target class $c$:
\begin{equation}\label{eq:determinstic_attack}
    \underset{\vct \delta}{\operatorname{argmin}} \; K(f^{\theta}(\vct x), \vct e_c) \ , \quad s.t. \lVert \vct{\delta} \rVert_p < \epsilon \ ,
\end{equation}
where $f^\theta(\vct{x})$ denotes the feature vector parameterized with $\theta$, and $K$ the RBF kernel.
As mentioned above about probabilistic models, also the efficiency of attacks against deterministic models is affected by the choice of a proper target.
In the case of DUQ, the attack can be crafted more easily by targeting the class \emph{nearest} to the centroid.
Due to the deterministic nature of the considered models, we argue that in the absence of an adversarial training technique, it is quite easy for the attacker to craft the desired attack sample. 
Furthermore, the direct correspondence between the uncertainty measure and the distance to the closest centroid makes DUQ even less robust to attacks, since attacking the prediction also results in minimizing the uncertainty, with no additional perturbation required.

\subsection{Case Study: Semantic Segmentation}
\label{sect:atk_semantic_segm}
We have shown how attacks targeting the uncertainty measure can be formulated for standard classification problems.
Here we show how they can be extended to complex computer vision problems such as semantic segmentation, which can be seen as a multivariate classification problem, where a class label is assigned to each pixel. In this task, uncertainty is computed in a pixel-wise manner. To this aim, two commonly used metrics for aleatoric and epistemic uncertainty are the average prediction entropy and the prediction variance, respectively~\cite{kendall17-what}.

While recalling that it is common for segmented objects to present high uncertainty along the edges, we directly apply our attack formulation of Eq.~\ref{eq:mva} to semantic segmentation and, indeed, we find it challenging to decrease the uncertainty measure around the \emph{edges} of the segmented objects (see Sect.~\ref{sect:experiments_semantic}). 
We hypothesize this is due to the fact that the network is ``forced'' to abruptly change prediction around the edges, which are therefore inherently characterised by high uncertainty.
We, therefore, devised an application-specific attack to semantic segmentation. 
The underlying rationale is that a pixel closely surrounded by several pixels from different classes exhibits a correlation to each of such classes, whereas a pixel surrounded by a region mostly belonging to a single class exhibits a high correlation only with that specific class. 
Accordingly, we can force the network to predict a single, identical class for all the image pixels by minimizing the pixel-wise cross-entropy:
\begin{equation}\label{eq:semantic_segmentation}
    \underset{\vct \delta}{\operatorname{argmin}} \; -\sum_{\omega \in \Omega}  \log(f(\vct{x+\delta})_{\omega,c}) \ , \quad s.t. \lVert \vct{\delta} \rVert_p < \epsilon \ ,
\end{equation}
where $c$ denotes the index of the target class, $\Omega$ the set of pixels, and $f(\vct{x+\delta})_\omega$ the predicted probability vector for the pixel $\omega$.
The target class $c$ should be chosen as the one that minimizes the uncertainty. 
To this aim, as a rule of thumb, one can choose the \textit{most representative class}, i.e., the class corresponding to the majority of pixels in the predicted segmentation map. 
The above criterion presents multiple advantages. 
First, the attacker will be trivially required to flip the smallest number of pixels, as the majority of them are already assigned to the target class.
Secondly, considering the strong overall correlation induced by massive occurrences of pixels of the most representative class $c$ on the image, a pixel of a different class can be misled towards $c$ with much more ease compared to a different and less impactful class. 
We refer to this attack with the name \textit{Uniform Segmentation Target Attack (UST)}.

\section{Experimental Analysis}\label{sect:experiments}

We empirically evaluated the proposed O-Attack against several UQ techniques both in classification and semantic segmentation tasks, under two different operational scenarios: the traditional setting of \textit{independent and identically distributed} (IID) data, which is practically implemented using training and testing data from the same data set, and the case of \textit{out-of-distribution} (OOD) data, which was simulated using different data sets for training and for testing.

\subsection{Experimental Setup}\label{sect:experimental_setup}

\myparagraph{Data sets} 
We used CIFAR-10 for IID experiments, whereas for OOD experiments we used CIFAR-10 for training and CIFAR-100 for testing. 
To evaluate the performance under the OOD setting, we used accuracy-rejection curves evaluated on a mixed testing set made up of 600 CIFAR-10 samples and 900 CIFAR-100 samples.

We further assessed the O-Attack in a semantic segmentation task, on the PASCAL VOC data set~\cite{Everingham15}. 

\myparagraph{UQ techniques and models} 
We considered four different UQ methods: \textbf{MC dropout}~\cite{GalG16} (implemented both in ad hoc and post hoc fashion), \textbf{Deep Ensemble}~\cite{Lakshminarayanan17} and \textbf{DUQ}~\cite{van2020uncertainty}. 
We also considered three DNN architectures to implement the models: ResNet18, ResNet34 and Resnet50. 
We trained 9 different versions of ResNet34 and Resnet50 and 10 versions of ResNet18: one baseline version for the post hoc dropout, 3 versions with ad hoc dropout (using a dropout rate $\, \phi\, \in\, [0.1, 0.3, 0.5]$, and five classic ResNet's for constructing a deep ensemble. 
For models including MC dropout-based architectures, we added the dropout rates after each convolutional and linear layer, thus obtaining a probability distribution over each weight. 
For ResNet18, we trained an additional network used as a feature extractor for DUQ.
For the semantic segmentation task, we used the pre-trained Torch implementation of a Fully Convolutional Network (FCN)~\cite{LongSD15}. 
We then applied post hoc dropout with a dropout rate of $0.1$ after each block of four convolutions, since a too high randomization may induce prediction deterioration when using injected dropout~\cite{Ledda2023}.

\myparagraph{Attack implementation}
We based the implementation of our attack (see Sect.~\ref{sect:uq_under_atk}) on the PGD attack with \ellinf norm, using $150$ iterations, MC samples size of $30$ and step size of $2\cdot 10^{-3}$ for the case of probabilistic UQ methods, and $10$ iterations with a step size of $1\cdot 10^{-3}$ for the deterministic method DUQ.
For the MVA attack of Eq.~\ref{eq:mva}, we minimize the logarithm of the variance to attain better performances. 
We implemented the attacks on CIFAR-10 with $\epsilon$ ranging from $1/255$ to $8/255$. 
This allowed us to plot the associated security evaluation curves showing how the uncertainty measure changes as a function of $\epsilon$. 
For semantic segmentation, we still use the PGD attack with $100$ iterations, a step size of $1\cdot 10^{-3}$, set $\epsilon$ to $2/255$ and MC sample size of $20$. 

\myparagraph{Uncertainty measures}
To attack probabilistic UQ methods, we use MC sample size of $100$ (for both classification and segmentation) to estimate the \textbf{predictive variance} and the \textbf{entropy} as measures of \emph{epistemic} and \emph{aleatoric} uncertainty, respectively. 
To attack DUQ, we use the distance from the closest centroid to measure epistemic uncertainty.

\subsection{Experimental Results}
\label{sect:results}
We first present and discuss the results attained by attacking \emph{probabilistic} UQ, for both IID and OOD data, then the ones attained for the \emph{deterministic} DUQ method, and finally the results related to semantic segmentation.

\myparagraph{Probabilistic UQ methods, IID setting}
\begin{figure}[h!]
    \centering
    \includegraphics[width=\linewidth]{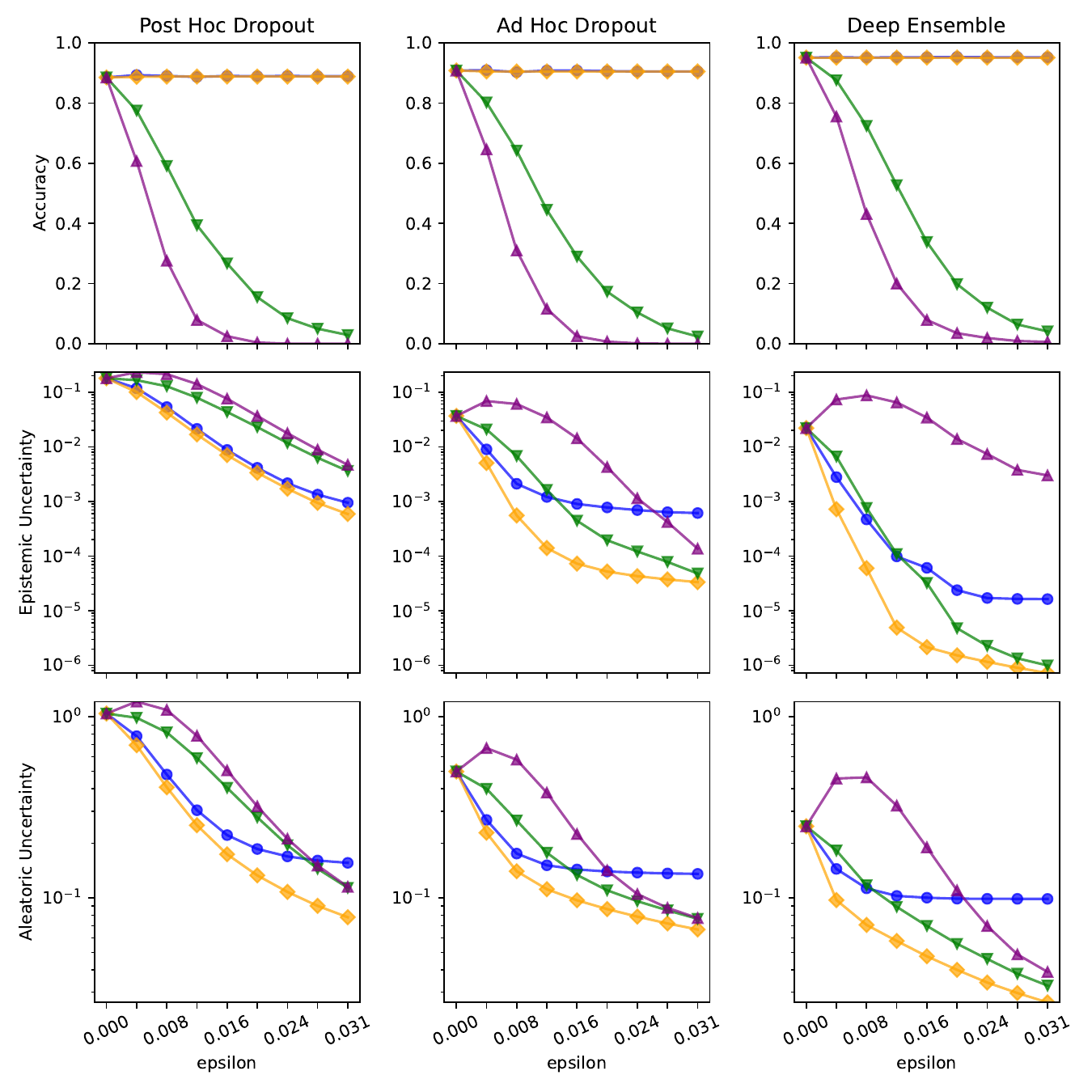}
    \includegraphics[width=\linewidth]{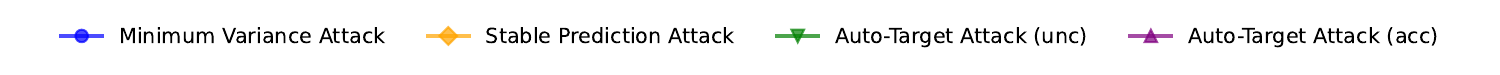}
    \caption{Behaviour of classification accuracy, aleatoric and epistemic uncertainty on CIFAR-10 under an IID setup, using a ResNet18 model with MC-dropout (with a dropout rate of $0.3$) and Deep Ensembles, under different attacks, as a function of $\epsilon$.
    More architectures and dropout rates are present in the supplementary material.
    }
\label{fig:seceval}
\end{figure}
Fig.~\ref{fig:seceval} shows the results of the experiments conducted on CIFAR-10, in the IID setup, for all the considered probabilistic UQ methods.
The Minimum Variance Attack (MVA) and Stabilizing Attack (STAB) are conceived to minimize the uncertainty measure. 
MVA focuses on minimizing epistemic uncertainty, whereas STAB focuses on the predictive measure, thus minimizing both epistemic and aleatoric uncertainty. 
Interestingly, although not surprisingly, we can see that STAB turned out to be more effective in minimizing aleatoric uncertainty. 
In fact, pushing towards more stable predictions ultimately yields the double effect of increasing the target class probability and minimizing the entropy. 
However, for both attacks, the clean accuracy does not suffer any decline.

Whereas ATA is initially less efficient than STAB (since it attempts to induce misclassifications) both techniques stabilize as the attack proceeds. 
Such ATA behavior is caused by the initial warm-up phase described in Sect.~\ref{sect:atk_prob_models}, where the predictions necessarily cross the boundary before being uniformly pushed towards the same class. 
However, there are still some differences between the two techniques, indicating that ATA does not necessarily converge to STAB's performances for $\epsilon=8/255$ (e.g., on post hoc dropout).

For what concerns the comparison between UQ methods based on ad hoc and post hoc dropout, we did not observe any significant difference. 
From a broader perspective, MVA attacks appear to better fit post hoc dropout, whereas ATA seems more effective for ad hoc dropout. 
However, in both cases, STAB outperforms MVA and ATA for both aleatoric and epistemic uncertainty.
Overall, post hoc dropout attains a higher starting variance, which results in more difficulties in zeroing the uncertainty.

Besides being Deep Ensembles widely recognized as highly accurate techniques, in our experiments, we notice a conflicting trend. 
Starting from comparable uncertainty levels with respect to ad hoc dropout, we notice a considerable decline in both aleatoric and epistemic uncertainty. 
In fact, all the attacks easily reduce variance to the order of magnitude of $10^{-6}$ with a perturbation of $\epsilon=8/255$, conversely to the ad hoc dropout, which attains an order of magnitude of $10^{-4}$. 

As shown in Fig.~\ref{fig:seceval} and already stated in~\cite{Carlini2017}, a ``standard'' attack aiming to cause wrong predictions (denoted as ATA \textit{(acc)}) can also reduce uncertainty. 
However, by looking at Fig.~\ref{fig:seceval}, we find out that the criterion for choosing the \textit{best} adversarial example at each iteration is crucial. 
Indeed, in a traditional set-up, when we find an adversarial example fooling the prediction (i.e., misclassified by the model), we consider it a ``success'' and then save it. 
Nevertheless, this strategy is sub-optimal when an attacker is interested in evading the uncertainty measure. Indeed, we observe a first warm-up phase where the sample's uncertainty increases and then a stabilization where it consistently decreases (as hypothesized in Sect~\ref{sect:atk_prob_models}). 
Conversely, when always saving the sample with lower uncertainty, the uncertainty measures decrease consistently, as expected in this setting.

\myparagraph{Probabilistic UQ methods, OOD setting}
\begin{figure}[t]
    \centering
    \includegraphics[width=0.45\textwidth]{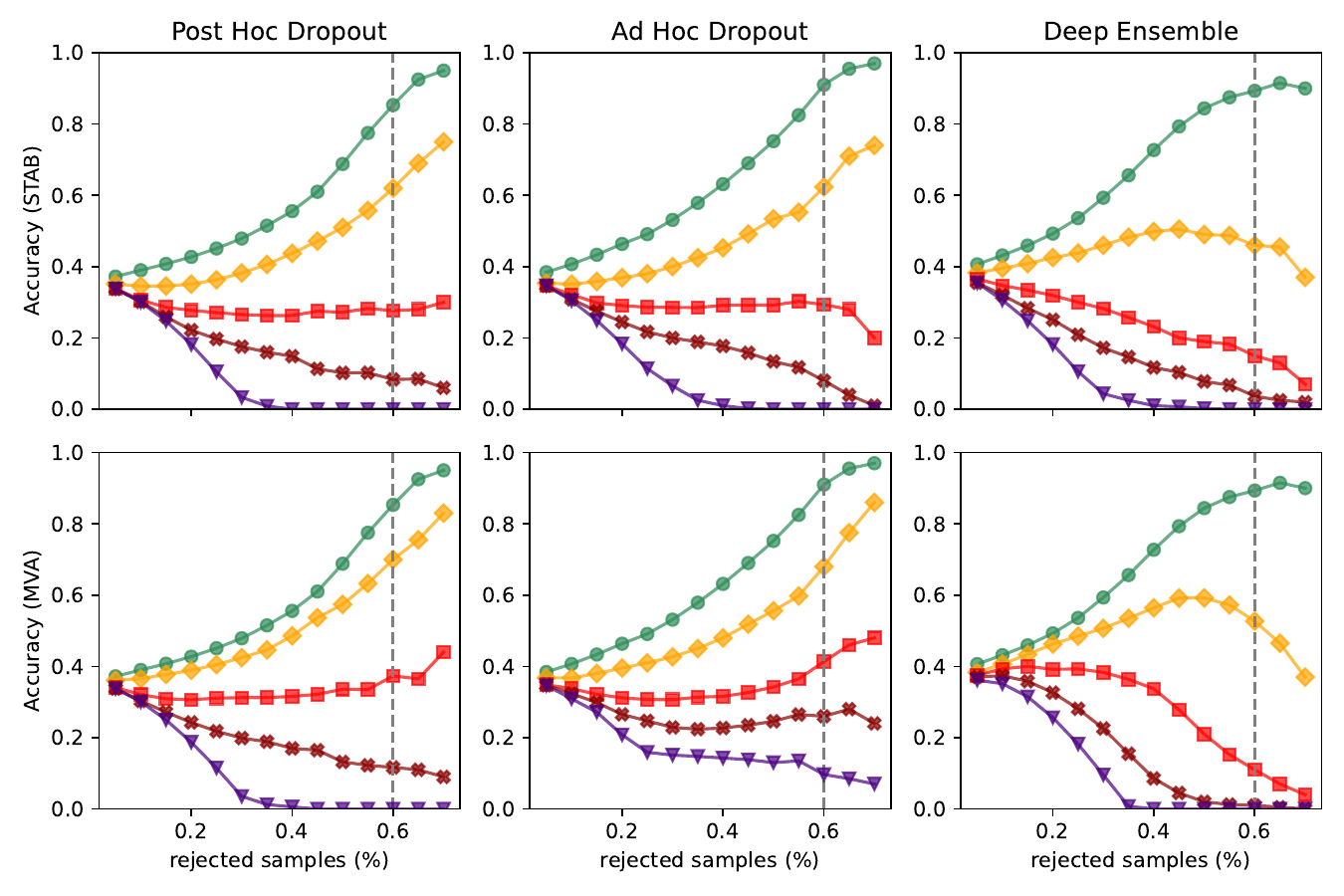} 
    \includegraphics[width=0.45\textwidth]{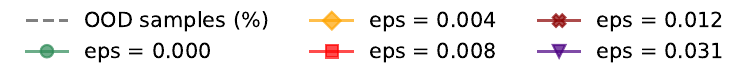} 
    \caption{
    Accuracy-rejection curves of a ResNet18 model under the STAB and MVA attacks against MC-dropout (with a dropout rate of $0.3$) and against Deep Ensembles, as a function of $\epsilon$, in a OOD setting simulated with a mixture of 600 CIFAR-10 images and 900 CIFAR-100 testing images.}
    \label{fig:rejection}
\end{figure}
We focused on the STAB and MVA attacks applied to post hoc dropout, ad hoc dropout and Deep Ensemble.
Fig.~\ref{fig:rejection} shows the corresponding accuracy--rejection curves. 
The green line, for $\epsilon=0$, shows that all UQ methods exhibit an adequate capability of detecting OOD samples. 
However, as the perturbation $\epsilon$ for OOD samples increases, their effectiveness decreases, up to a point where they start rejecting IID samples before OOD ones, which indicates that the estimated uncertainty is higher for OOD than for IID samples: this is just the opposite behaviour to the desired one (i.e., indicates an attack success). 

The above results clearly show that the considered UQ methods, including Deep Ensembles, are vulnerable to adversarial attacks, also in the presence of OOD samples. 
We also point out that for ad hoc dropout, the MVA attack turned out to be less effective than STAB, which, with $\epsilon=8/255$, completely breaks the other techniques.

We finally argue that the robustness of Deep Ensembles could be improved by increasing the ensemble size (which was set to 5 in our experiments), although at the expense of an increase in processing cost.

\myparagraph{Deterministic UQ methods}
\begin{figure}
    \centering
    \includegraphics[width=\linewidth]{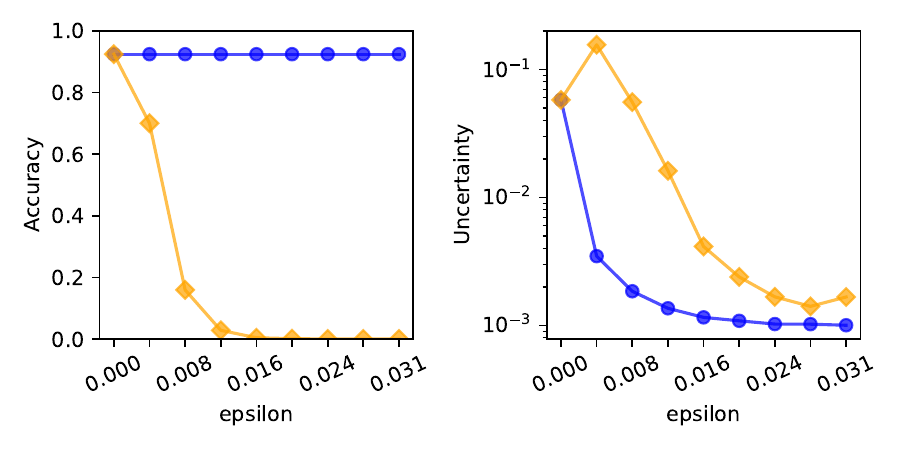}
    \includegraphics[width=\linewidth]{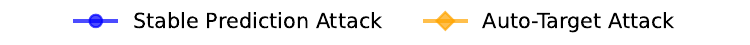}
    \caption{Classification accuracy and uncertainty of a ResNet18 as a feature extractor for the DUQ method on CIFAR10 in the IID setting, under two different attacks, as a function of $\epsilon$.}
    \label{fig:seceval_duq}
\end{figure}
\begin{figure}[b]
    \centering
    \includegraphics[width=0.7\linewidth]{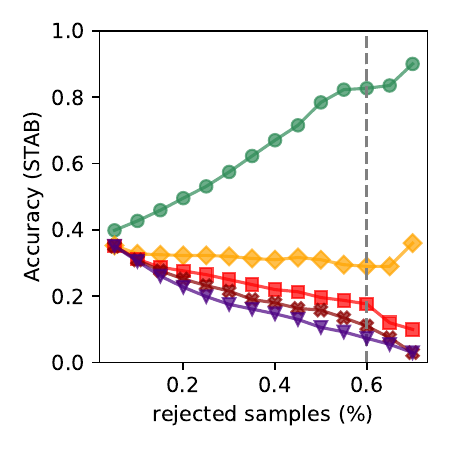}
    \includegraphics[width=\linewidth]{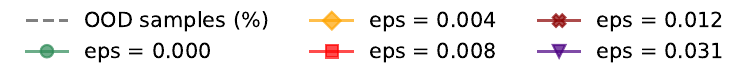}
    \caption{Accuracy-rejection curves attained in a OOD setting (see Sect.~\ref{sect:experimental_setup}) by a ResNet18 model using the DUQ method, under the STAB attack.}
    \label{fig:rejectioin_duq}
\end{figure}
\begin{figure*}[h!]
    \centering 
    \includegraphics[width=0.49\textwidth]{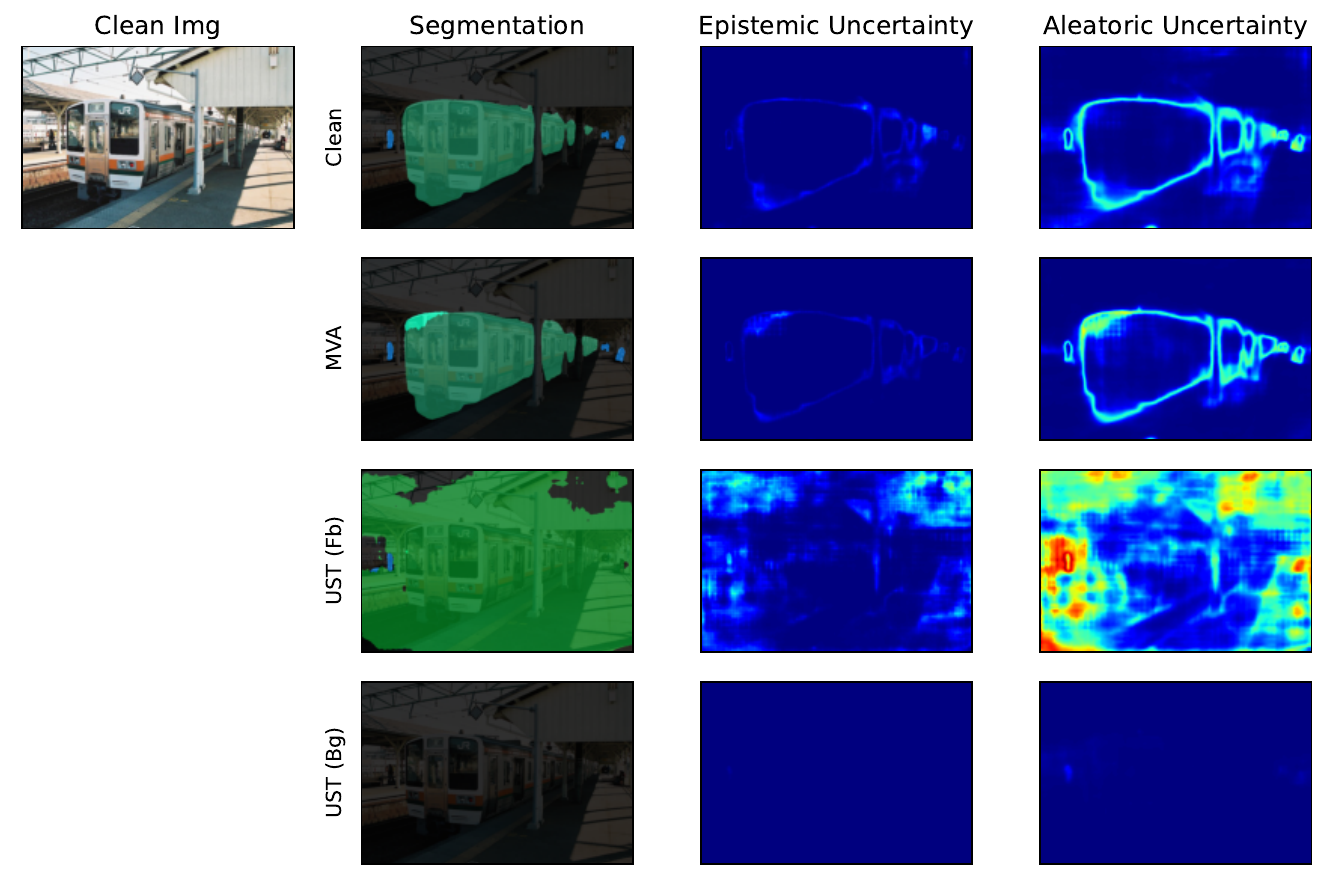}
    \includegraphics[width=0.49\textwidth]{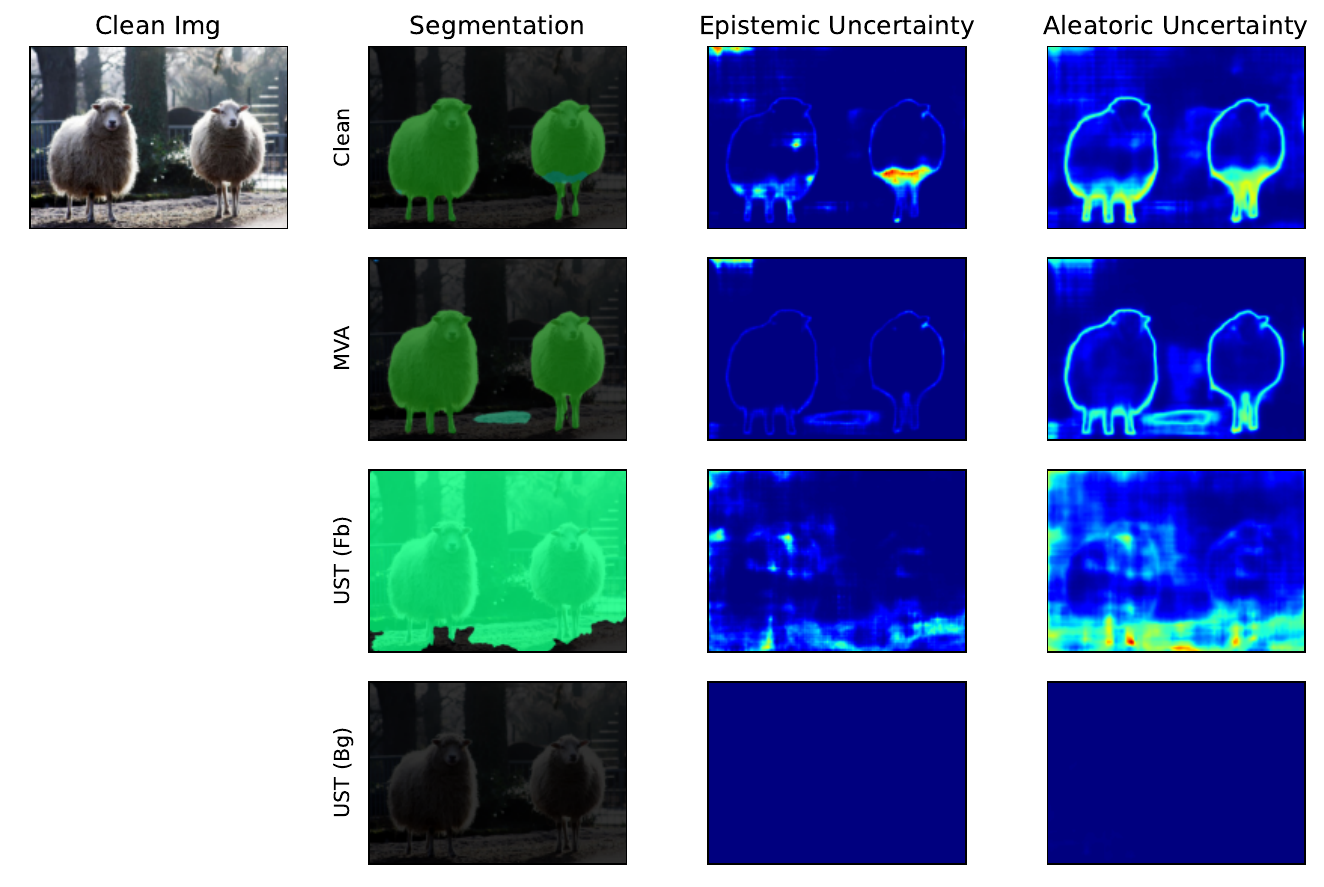} 
    \caption{Two examples of attacks against UQ in a semantic segmentation task. 
    In each of the two groups of columns, from left to right: (i) the original clean image, (ii) the predicted segmentation maps, and the corresponding epistemic (iii) and aleatoric (iv) uncertainty maps.
    In the rows, from top to bottom: results obtained under normal operating conditions (with no attacks), and under the MVA, UST (Fb), and UST (Bg) attacks.}
    \label{fig:segmentation}
\end{figure*}
In Fig.~\ref{fig:seceval_duq} and Fig.~\ref{fig:rejectioin_duq} we can see the results for IID and OOD (respectively) experiments using DUQ.
Since deterministic methods do not perform MC sampling, attacks against them can be designed and implemented more easily. 
This leads to lower robustness for attacks targeting both uncertainty and predictions (where, as opposed to probabilistic attacks, no trade-off is needed).
Nevertheless, more interesting behaviors can be observed when DUQ is used in the case of OOD samples, as seen from Fig.~\ref{fig:rejectioin_duq}.
In this scenario, even small perturbations quickly deteriorate the quality of the uncertainty measure. Still, for larger perturbations, the accuracy does not drop to zero: such behavior may indicate that deterministic methods assign larger uncertainty values to OOD samples, making it challenging to get the perturbed samples very close to a target centroid.

\myparagraph{Semantic segmentation}\label{sect:experiments_semantic}
We finally show in Fig.~\ref{fig:segmentation} the results obtained when attacking UQ methods used for a semantic segmentation task.
Using a clean image as input, we see that the considered model is not very accurate in correctly segmenting the whole object. Nevertheless, high uncertainty values are correctly assigned to regions where segmentation errors occur, corresponding to the object edges and to missing objects. 
MVA, albeit reducing the overall epistemic uncertainty, is less effective in reducing the uncertainty on the edges. 
On the other hand, the attack aimed at obtaining a uniform segmentation map whose target is the most representative class (usually, the ``background'' class), which we refer to as UST (Bg) for convenience, turns out to be effective in reducing both the epistemic and the aleatoric uncertainty for each pixel. 
However, attacks aimed at evading the predictions using a similar strategy, i.e., assigning a wrong label (referred to as UST (Fb), \ie ``Full Break''), did not achieve a similar reduction in uncertainty, despite they evaded a large region of the image.

\section{Conclusions and Future Work}

In this work, we first proposed and modeled adversarial attacks against UQ techniques used by ML predictors, aimed at producing \emph{wrong uncertainty estimates}, regardless of the correctness of the prediction.
We formally defined a taxonomy and a threat model 
and implemented several possible attacks against different UQ techniques, both in classification and in semantic segmentation tasks.

From our preliminary results on classification tasks we can draw the following conclusions:
Generally speaking, \textbf{UQ techniques are not robust to adversarial attacks}: they can be easily manipulated using attacks specifically crafted to evade the uncertainty measure. 
Surprisingly, Deep Ensemble turned out to be the less robust UQ technique against adversarial attacks targeting uncertainty. 
On the other hand, MC dropout tends to be the most robust among the analyzed methods (as we can see from the experiments on OOD data). 
Our preliminary example on semantic segmentation shows that attacks against UQ methods can be effective also in other, more complex CV tasks.

We finally point out the following directions for future work: (i) implementing and investigating under-confidence attacks (U-attacks); (ii) exploring the proposed attacks against a wider range of UQ methods; (iii) analyzing black-box attacks;
and (iv) exploring the attack transferability between different UQ methods; (v) investigating adversarial training and other robust defense techniques to counter attacks against UQ.
\section*{Acknowledgments}

This work has been supported by the European Union's Horizon Europe research and innovation program under the project ELSA, grant agreement No 101070617; by Fondazione di Sardegna under the project ``TrustML: Towards Machine Learning that Humans Can Trust’’, CUP: F73C22001320007; and by project SERICS (PE00000014) under the NRRP MUR program funded by the EU - NGEU.

Emanuele Ledda, Daniele Angioni, and Giorgio Piras are affiliated with the Italian National Ph.D. in Artificial Intelligence, Sapienza University of Rome. 
They also acknowledge the cooperation with and support from the Pattern Recognition and Applications Laboratory of the University of Cagliari. 

{\small
\bibliographystyle{abbrv} 
\bibliography{egbib}
}

\end{document}